# SFI-Swin: Symmetric Face Inpainting with Swin Transformer by Distinctly Learning Face Components Distributions

MohammadReza Naderi[1*], MohammadHossein Givkashi[1*], Nader Karimi[1], Shahram Shirani[2], Shadrokh Samavi[1,2,3]

[1]Department of Electrical and Computer Engineering, Isfahan University of Technology, 84156-83111, Iran
[2]Department of Electrical and Computer Engineering, McMaster University, L8S 4L8, Canada
[3]Computer Science Department, Seattle University, Seattle 98122 USA

**Abstract**

Image inpainting consists of filling holes or missing parts of an image. Inpainting face images with symmetric characteristics is more challenging than inpainting a natural scene. None of the powerful existing models can fill out the missing parts of an image while considering the symmetry and homogeneity of the picture. Moreover, the metrics that assess a repaired face image quality cannot measure the preservation of symmetry between the rebuilt and existing parts of a face. In this paper, we intend to solve the symmetry problem in the face inpainting task by using multiple discriminators that check each face organ's reality separately and a transformer-based network. We also propose "symmetry concentration score" as a new metric for measuring the symmetry of a repaired face image. The quantitative and qualitative results show the superiority of our proposed method compared to some of the recently proposed algorithms in terms of the reality, symmetry, and homogeneity of the inpainted parts. The code for the proposed method is available at https://github.com/mohammadrezanaderi4/SFI-Swin

## 1. Introduction

Removing objects from an image or filling in holes is a typical application of computer vision. With the image inpainting technique, it is possible to either fill in empty regions or remove a few elements from the image. New deep learning models with convolutional neural networks or transformer models are intended to produce realistic-looking inpainted images. Face inpainting is a subset of image inpainting. Its purpose is to fill the missing regions of a face image. Two major concerns should be considered carefully during the inpainting of missing parts of a face: First, the inpainted regions must be homogeneous with the other parts of the face and highly correlated to the available surrounding areas of the input image. Second, facial symmetry must be preserved between the left and right sides. Many inpainting methods have been proposed, and some achieved excellent results in repairing missing areas of natural images. But almost all have difficulty repairing a face image symmetrically and homogeneously. This shortcoming is because the network losses do not convey a general understanding of the facial features to the generator. To further illustrate the main issues of previous works, we will discuss the effect of usual losses that have been used in references [1] to [6].

The loss functions that are mainly used in inpainting are pixel-wise, adversarial, feature-matching, and perceptual loss. We will discuss the effect of each loss on the model training in this section.

**1.1 Pixel-Wise Loss:** As shown in Equation 1, Pixel-wise loss is computed between the inpainted image and ground truth. Its goal is to lead the model to inpaint the missing regions similar to ground truth by considering available parts of the face. However, the available regions cannot completely describe the missing parts of the image. Therefore, this loss only can lead the inpainting network to understand the low-level features of the missing parts.

$$\mathcal{L}_{PW}(x, \hat{x}) = \|x - \hat{x}\|_{1 \text{ or } 2} \qquad (1)$$

In Equation 1, $x$ is the inpainted image, $\hat{x}$ is the ground truth, and $\| \|_{1 \text{ or } 2}$ stands for L1 or L2 norm computation.

---
[*] The first two authors contributed equally to this work.

**1.2 Adversarial and Feature Matching Losses:** The adversarial loss [7] (Equations 2, 3, 4) attempts to check the reality of an inpainted image based on the distribution of ground truths and generated images using an extra network called a discriminator. In Equations 2, 3, and 4, $x, \hat{x}, D, G, \varepsilon$, and $\theta$, represent the inpainted image, ground truth, discriminator, generator, parameters of the discriminator, and parameters of the generator. Also, $sg$ is the stop gradient, indicating that the backpropagated gradient stops when it reaches specific parameters. Finally, the feature matching loss (Equation 5) is computed between the features extracted from the discriminator's middle layers for the inpainted image and ground truth. In Equation 5, $MFD$ presents middle features of the discriminator. Adversarial and feature matching losses came from the idea that although we cannot reconstruct the missing regions exactly similar to ground truth, at least we can be sure that the inpainted regions look realistic.

$$\mathcal{L}_D = -E_x[\log D_\varepsilon(x)] - E_x[\log(1 - D_\varepsilon(\hat{x}))] \qquad (2)$$

$$\mathcal{L}_G = -E_x[\log D_\varepsilon(\hat{x})] \qquad (3)$$

$$\mathcal{L}_{Adv} = sg_\theta(\mathcal{L}_D) + sg_\varepsilon(\mathcal{L}_G) \qquad (4)$$

$$\mathcal{L}_{FM}(x, \hat{x}) = \| MFD_\varepsilon(x) - MFD_\varepsilon(\hat{x}) \|_{1 \text{ or } 2} \qquad (5)$$

The architecture of the discriminator is patch-based. The discriminator tests the reality of the patches of the whole image [7].

**1.3 Perceptual Loss:** The perceptual loss is shown in Equation 6. This loss is computed using the encoder of a pre-trained segmentation network to compare the high-level features of the generated and ground truth images. In Equation 6, $x, \hat{x}, \phi$, present inpainted image, ground truth, and encoder of a pre-trained segmentation network, respectively. This loss mostly considered high-level features such as edges in the image.

$$\mathcal{L}_{PL}(x, \hat{x}) = \| \phi(x) - \phi(\hat{x}) \|_{1 \text{ or } 2} \qquad (6)$$

The adversarial, feature matching, and perceptual losses mainly focus on the reality of the patches and edges' smoothness. Therefore, these losses do not represent the importance of the homogeneity and symmetry of the image. The perceptual loss causes the inpainting network to occasionally prefer the reality of the patches over the validity of the whole picture. In other words, due to the pressure of the mentioned losses, the generator sacrifices the symmetry of the face to get more realistic patches. This patch-based behavior is one of the apparent drawbacks of available image synthesizing and image inpainting methods. The mentioned shortcoming motivates us to add a new term to the loss function of image inpainting networks to consider symmetry and global features of each part of the face. We can get more realistic faces by assessing an image's symmetry and global facial features. Meanwhile, we use a base inpainting network that balances global and patch-based losses to get the maximum benefit from this additional loss.

A large receptive field is used in Swin transformer blocks [8]. We used Swin transformer blocks in this paper and applied a new loss focusing on distinct facial feature distribution. Hence, we increase the inpainting model's concentration on the symmetry problem. Furthermore, current metrics cannot measure the symmetry of the inpainted face. Therefore, we propose a new metric to resolve this shortcoming.

Our main contributions are as follows:

- Proposing a homogeneity-aware loss to solve the heterogeneity and asymmetricity problems in inpainting methods.
- Using transformer base architecture to achieve a wider receptive field and balance the usage of global and local features and input and output image gradients.
- Increasing the generalizability of the inpainting model by upgrading its understanding of the semantic face parts using separate discriminators for each part of the face.
- Proposing a new metric, symmetry concentration score (SCS), to assess the symmetry and homogeneity of facial organs.

The structure of this paper is as follows: In section 2, we will present some previous relevant methods. Then, section 3 offers the proposed method. Finally, experimental results are detailed in section 4, and the conclusion is presented in section 5.

## 2. Related Works

Several methods for image inpainting have been proposed, including deep generative base and transformer base methods. This section will discuss these two types of methods in more detail.

### 2.1 Deep Generative Methods

In recent years, the generative adversarial network has been used for image completion and inpainting [9]–[12]. In [1], a new approach has been proposed based on conditional and unconditional generative networks. The authors present new metrics for measuring image completeness. These metrics are based on the linear separability of features in a feature space. These metrics demonstrate a measure of the perceptual fidelity of the inpainted image compared to the real image. A method for generating high-quality inpainted image based on aggregated contextual transformations was presented in [2]. Authors of [2] also employed distant contexts, not in the masked region's neighborhood. Inpainting methods are evaluated on different tasks, such as removing logos, editing faces, and removing objects [13]–[16]. In [3], the authors focused on a large missing region and proposed an adversarial method for image inpainting. The model can generate visually realistic images for contiguous and separated large missing areas. Also, a new loss was proposed to measure the non-local correlation between patches and help the model get a better inference result.

A method is proposed in [4] that uses edges and masked images to generate inpainted images. They used an edge generator that completed edges in missing regions and a completion network that used an edge map and input image to generate the result. With this approach, they achieved a good result with a focus on the edges of the image. In [17], they introduced a method for human body completion. They used three steps in their work: prior encoding, segmentation completion, and texture completion. A segmentation map can help capture important information from the human body. They also designed a memory module, a dictionary for storing learned latent vectors. These segmentation maps are typically used to segment each part of the human body and assist in extracting high-level information. As a result, it helps the model to learn better. Multi-scale structure discriminators have been used to generate segmentation maps with good information, so the results were interesting in human body completion. In [18], the authors used a generative adversarial network that contains dilated convolution for increasing receptive fields. They tried to achieve a reasonable structure without blurriness in the output. Further, they developed a new self-guided regression loss technique for enhancing semantic features and concentrating on uncertain areas. When a model fills holes in an image, it's essential to generate a good result and have color consistency in all pictures.

In [19], the authors proposed a method for image inpainting with attention to color consistency in images. The output was reliable without any artifacts, and the stable colors in different parts of the images provided the ability to create realistic images. In [5], the authors used gated convolutions in the model to solve the problem of using vanilla convolutions that understand the pixels differently. They also proposed a new loss function named SN-PatchGAN that helps the model generate a high-quality and flexible image compared to previous methods.

### 2.2 Transformer Methods

Since transformers are highly effective for tasks related to natural language processing, researchers have used them in vision tasks. A transformer has a global view and uses an attention mechanism to extract information [20–23]. In [24], a transformer was used for image completion. They use the global view of the transformer and achieve better results for large masks than other methods. Recently a combination of transformer and convolution has been used. In [25], the authors proposed a model containing a transformer and convolution for image inpainting. The global view of the transformer and feature extraction in convolution helped the model generate good results in image completion. In [26], a novel inpainting framework has been proposed for high-resolution images. They also proposed modifying transformer blocks to stabilize the training of large masks.

With the long-range interaction modeling in the transformer, the model generated high-fidelity images. Therefore, the architecture can understand critical dependencies between different regions on the image with attention-based models. In [27], they proposed an inpainting method that used a transformer and CNN model. The model gets additional input that contains lines and edges on the masked image. As a result, the model can reconstruct edges in a better way. With this approach, the proposed model can learn to reconstruct masked images focusing on the edges and lines. In [28], the blind face inpainting method has been proposed. They used two stages for the blind face inpainting process due to the difficulty in detecting masks of different shapes and sizes, as well as the difficulty of restoring masked images that are realistic. They used a transformer model during the first stage to detect corrupted regions. In the next step, a network has been proposed for restoring features at different levels in a hierarchical manner, thereby producing semantically coherent content based on unmasked regions of the face.

# 3. Proposed method

Our proposed method is called Symmetric Face Inpainting with Swin Transformer (SFI-Swin). We will discuss our method from two perspectives. First, we will discuss the generator architecture based on a transformer in subsection 3.1. Then in subsection 3.2, we will concentrate on the loss functions and propose a loss that focuses on the symmetry and homogeneity of the face features. We use six additional discriminators with the same architecture to compute this loss.

### 3.1 Generator architecture

As discussed in [6], using a generator architecture with a wider receptive field produces more homogeneous outputs while considering the entire facial features. Thus, they added fast Fourier convolution layers [29] to their models to make a generator with a wider receptive field. Although their work seems to be effective, it sometimes fails to make a balance between using global and local features [6]. We use the Swin-Unet [30] architecture as our generator to create a balance between local and global features. Swin-Unet has a large receptive field because of its self-attention mechanism and could balance local and global feature usage well. As discussed in [31], skip connections downgrade the image inpainting ability of the generator model because it allows the generator to copy the available parts of the input to the output.

Skip connections prevent the model from constructing the whole available and missed parts of the input based on high-level features that are extracted in the middle of the generator. Such a network results in totally heterogeneous face features in output. Therefore, we omit the skip connections from the Swin-Unet model, and our final generator architecture is shown in Fig. 1. The patch discriminator [7] architecture is also shown in Fig. 1.

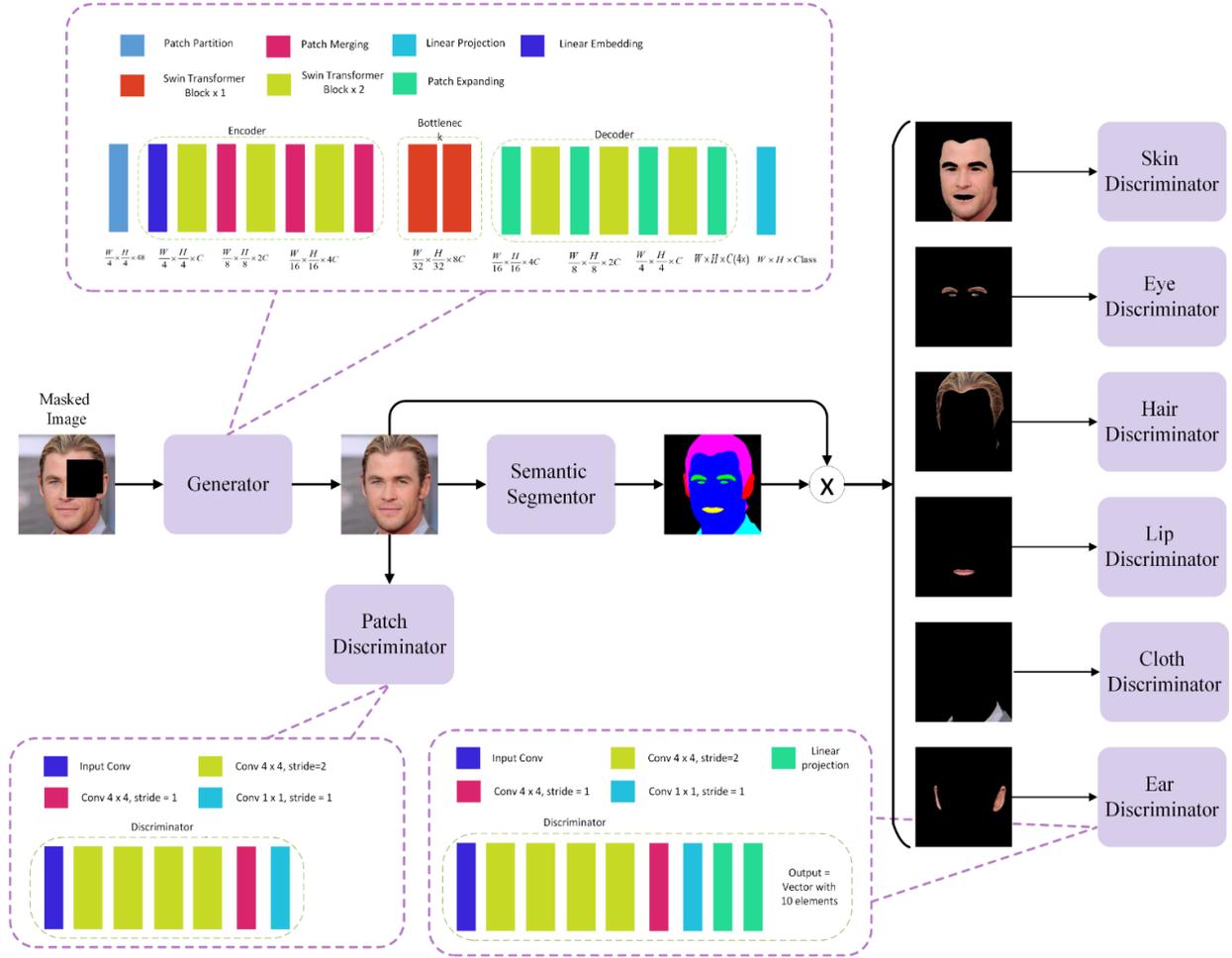

**Fig. 1.** Block diagram of our proposed method. First, the generator takes the masked image as input and attempts to inpaint it. Then the inpainted image is fed to the patch discriminator to check the overall reality of the patches. Meanwhile, the inpainted image is also fed into a semantic segmentation network [32] to separate the semantic parts of the face, such as eyes, and ears. The architecture of these six semantic discriminators is the same. In the next step, six distinct discriminators calculate the total realness of each semantic part of the face. Then, the generator parameters will be updated based on these seven discriminators and the pixel-wise loss gradient signals, which are backpropagated to the generator.

Using the Swin transformer [8] layers, which could balance the utilization of local and global features, leads to more homogeneous and realistic facial features. In addition, Swin transformer blocks do not add much computational complexity compared to convolutional models with a similar number of parameters. This behavior is because the Swin transformer computes the correlation between the patches in certain parts of the image.

### 3.2 Homogeneity Loss

In this section, we propose a new loss that we will use in addition to the losses that were discussed in Section 1. The idea behind this new loss is that we compute the realness of each part of the face compared to the distribution of that specific part in the whole dataset. This will oblige the generator to pay more attention to symmetric and global features. We consider six additional discriminators with the same architecture to compute the homogeneity loss, as shown in Figure 1. Each discriminator is for a specific part of the face, such as skin, eyes, lips, clothes, and hair. It is noteworthy that each of these parts is extracted from a face image using a pre-trained semantic segmentation module [32]. The segmentation module parameters are frozen during training. While the generator intends to inpaint the missed regions, computing the realness of each facial organ will conclude more symmetry in the output image. The architecture of these discriminators is designed to assess the total realness of a specific face.

This characteristic is unlike the patch discriminator, which does not completely understand the face image. One of the main features of the face is its symmetry which the related discriminator will consider. Using these semantic discriminators to check the realness of each face organ increases the generator's knowledge about different semantic parts of the image. Considering facial organs helps the network to repair portraits more realistically. The block diagram of SFI-Swin is shown in Figure 1.

For each of these discriminators and the corresponding facial organ, the adversarial and feature-matching loss is computed using Equations 4 and 5. Therefore the overall homogeneity loss could be presented as Equations 7 and 8.

$$\begin{cases} \mathcal{L}_{skin}(x,\hat{x}) = \mathcal{L}_{Adv}(x_{skin},\hat{x}_{skin}) + \mathcal{L}_{FM}(x_{skin},\hat{x}_{skin}) \\ \mathcal{L}_{eye}(x,\hat{x}) = \mathcal{L}_{Adv}(x_{eye},\hat{x}_{eye}) + \mathcal{L}_{FM}(x_{eye},\hat{x}_{eye}) \\ \mathcal{L}_{hair}(x,\hat{x}) = \mathcal{L}_{Adv}(x_{hair},\hat{x}_{hair}) + \mathcal{L}_{FM}(x_{hair},\hat{x}_{hair}) \\ \mathcal{L}_{lip}(x,\hat{x}) = \mathcal{L}_{Adv}(x_{lip},\hat{x}_{lip}) + \mathcal{L}_{FM}(x_{lip},\hat{x}_{lip}) \\ \mathcal{L}_{cloth}(x,\hat{x}) = \mathcal{L}_{Adv}(x_{cloth},\hat{x}_{cloth}) + \mathcal{L}_{FM}(x_{cloth},\hat{x}_{cloth}) \\ \mathcal{L}_{ear}(x,\hat{x}) = \mathcal{L}_{Adv}(x_{ear},\hat{x}_{ear}) + \mathcal{L}_{FM}(x_{ear},\hat{x}_{ear}) \end{cases} \quad (7)$$

$$\mathcal{L}_{HG}(x,\hat{x}) = \omega_1 \mathcal{L}_{skin}(x,\hat{x}) + \omega_2 \mathcal{L}_{eye}(x,\hat{x}) + \omega_3 \mathcal{L}_{hair}(x,\hat{x}) + \omega_4 \mathcal{L}_{lip}(x,\hat{x}) \\ + \omega_5 \mathcal{L}_{cloth}(x,\hat{x}) + \omega_6 \mathcal{L}_{ear}(x,\hat{x}) \quad (8)$$

where $\omega_1$, $\omega_2$, $\omega_3$, $\omega_4$, $\omega_5$, and $\omega_6$ are set to be 0.083, 0.25, 0.083, 0.25, 0.083, 0.083, and 0.25, respectively. The coefficients of eyes, ears, and lips are set three times greater than other parts of the face. This is because, during the experiments, we realized that the model is mostly incapable of maintaining the symmetry of these face organs.

## 4. Experimental results

In this section, we present the experiment setups and results accomplished using these setups.

### 4.1 Experiments setups

In the following, we will discuss the dataset, metrics, hyperparameters, and the total loss function we used to train our models.

**Data and metrics:** We use CelebAHQ [33] dataset. This dataset contains 28k images. We split the data to train, validate, and test, similar to [6]. Learned Perceptual Image Patch Similarity (LPIPS) [34] and Fr'echet inception distance (FID) [35] metrics are used to evaluate our proposed method performance. Compared to pixel-level L1 and L2 distances, LPIPS and FID are more suitable for measuring the performance of large masks when multiple natural completions are plausible. The experimentation pipeline is implemented using PyTorch [36].

**Training hyperparameters:** The learning rate of the generator is 0.001, and that of the seven discriminators is 0.0001. The discrimination process is more straightforward than the generation process, especially in the initial epochs of the training [6]. Hence, a lower initial learning rate for the discriminators allows the generator to converge faster during the initial epochs. This will balance the training procedures between the discriminators and the generator. This balance prevents the training procedure from collapsing. These learning rates also are optimized using a beam search algorithm to get the best results [6]. The batch size is also set to 20, and we train our model for 40 epochs using the Adam optimizer [37]. We trained our model with an NVIDIA GeForce RTX 3090 with 24GB RAM GPU.

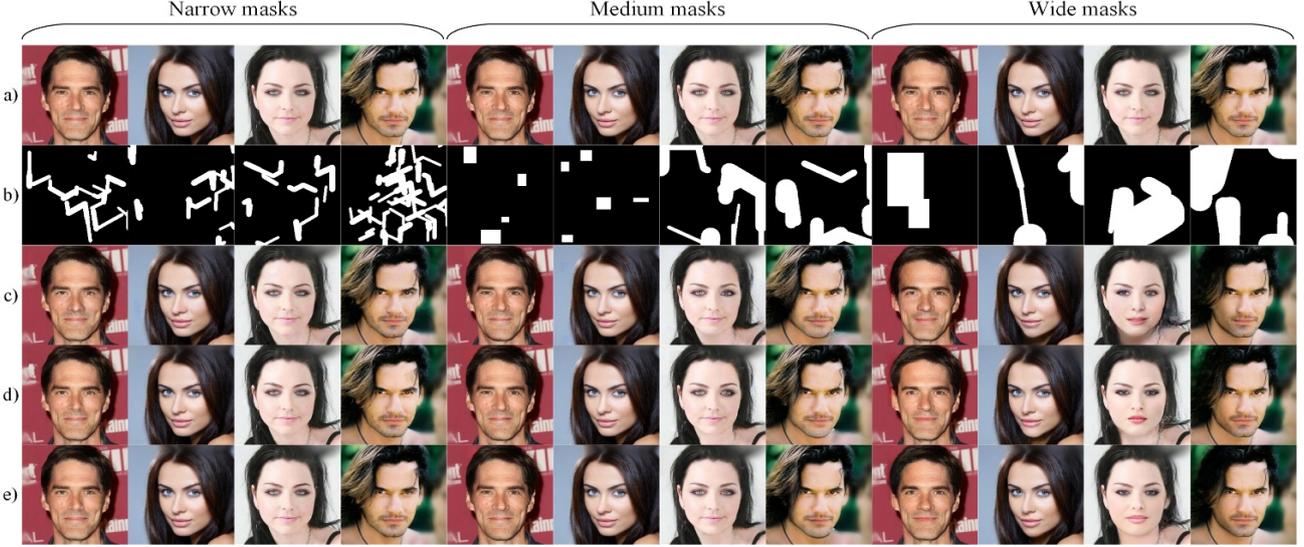

**Fig. 2.** a) Original images, b) masks, inpainted images by c) LaMa [6], d) Swin, and e) SFI-Swin.

**Loss function:** besides the proposed homogeneity loss, we used the same losses as [6]. Therefore, the total loss function of our model is shown in Equation 9.

$$\mathcal{L}(x, \hat{x}) = \alpha \mathcal{L}_{PW}(x, \hat{x}) + \beta \mathcal{L}_{Adv}(x, \hat{x}) + \gamma \mathcal{L}_{FM}(x, \hat{x}) + \delta \mathcal{L}_{HG}(x, \hat{x}) \qquad (9)$$

where $\alpha$, $\beta$, $\gamma$, and $\delta$ are set to 10, 10, 100, and 20. These values are based on our experiments to get the best results on the test data.

**4.2 Qualitative results:** Fig. 2 shows SFI-Swin performance to inpaint for different mask sizes, in comparison with [6]. To further illustrate our proposed method's capability to inpaint the face images symmetrically, we generate masks that omit one of the eyes and slowly grow around the omitted region to cover half of the face. We compare our method with [6], and the results are shown in Figure 3. Further, we randomly chose ten images from the test set. Then, we masked (eliminated) one eye and a K×K block of the image. We tested with K set to 16, 32, and 64.

We repaired the image with a missing eye and a K×K block. We then calculated the difference between the repaired eye while a K×K patch was masked with the inpainted image that only the eye was missing. We then find the mean of the absolute difference image and assign this value to the corresponding K×K block. The average absolute difference will be large if a block is essential to repair the missing eye. Finally, we built a heatmap to show the blocks that play an important role in the inpainting of the missing eye. The results are shown in Figure 4. As we can see, the non-missing eye and the chick area around the missing eye are the most important patches for reconstructing the missing area. We repeated this experiment for half of the face. The effects of each missing K×K patch on the inpaint eye and half face are shown in Figure 5. The lighter the color of the shown patch, the more effect it has on reconstructing the missing right-side eye.

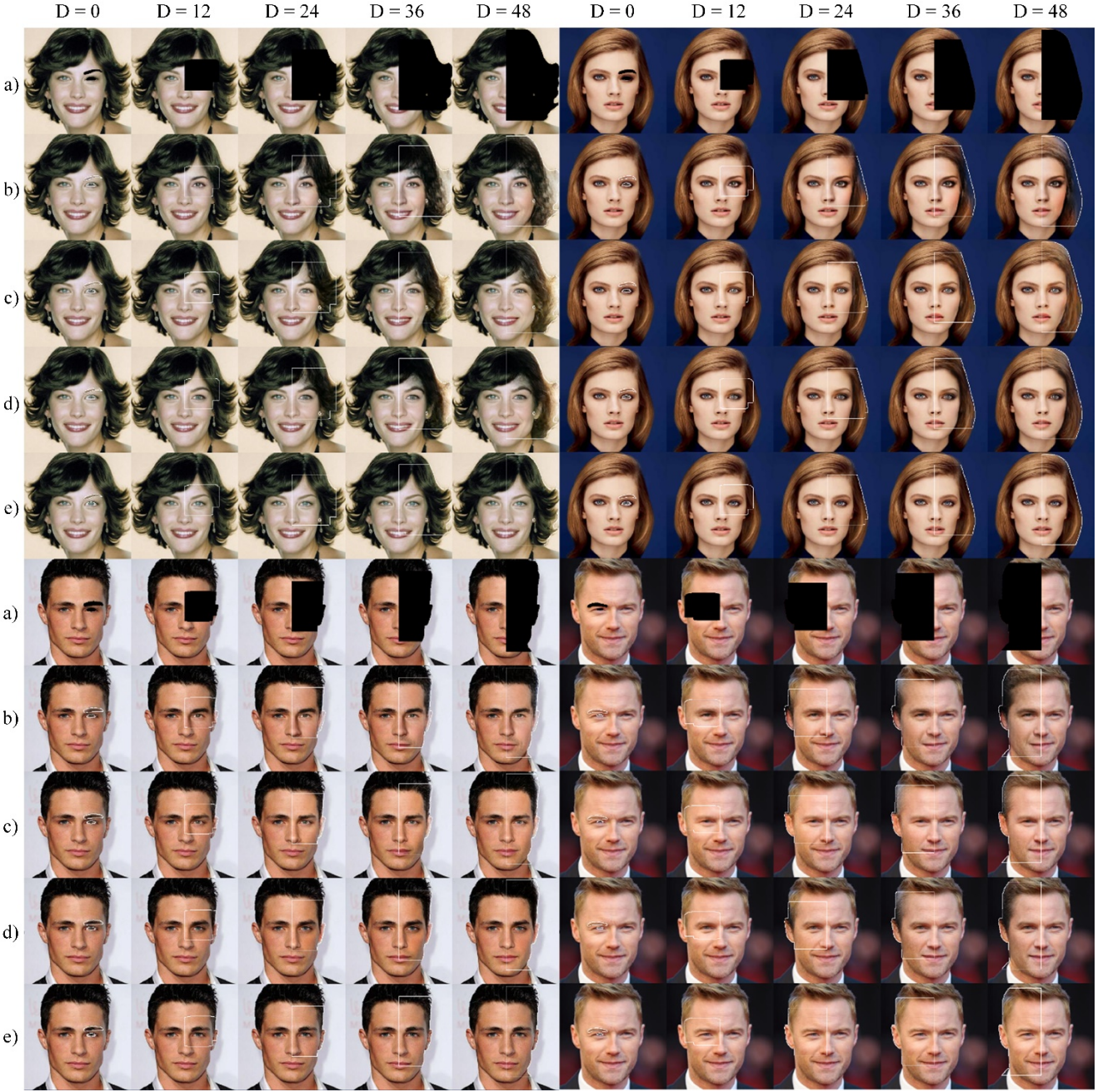

**Fig. 3.** Performances of b) LaMa, c) Swin, d) SFI-Swin in comparison with e) ground-truth to inpaint a) masked images starting from an eye and grown to cover half of the face. A white border is drawn to show the reconstructed region.

While inpainting a facial part from one side of the face, our method focuses on the same organ on the other side. This consideration causes symmetry in the repaired image.

**4.3 Quantitative results:** Table 1 shows the performance of our proposed method to repair the missing parts of the face compared with the best recently proposed methods in this field in terms of FID and LPIPS scores. Because previous works trained multiple models for different mask sizes, therefore to compare fairly, we show their performance based on the mask sizes which they used in training. However, our method follows the aggressive mask generation proposed by [6] that helps it handle narrow, medium, and wide masks simultaneously. As a result, our proposed methods, Swin and Swin with multiple semantic discriminators (SFI-Swin), achieved the best results in medium and wide mask inpainting.

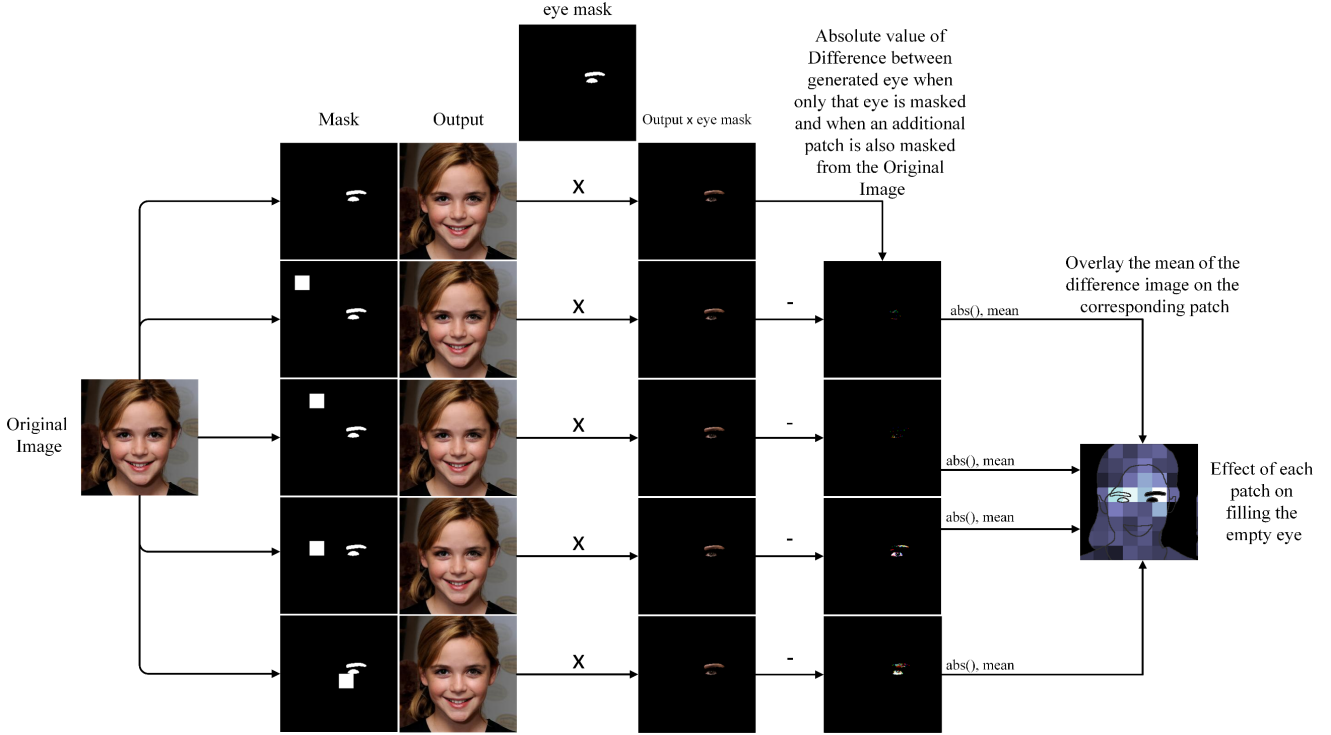

**Fig. 4.** Steps to compute symmetry concentration score (SCS). First, we mask an eye and a K×K patch of the face, and reconstruct the missing eye and the K×K patch. Then the absolute difference between the image with inpainted eye with the image with a missing K×K patch and the missing eye is computed. The difference shows the effect of that K×K patch on inpainting result of the missing eye. The effect of all K×K patches is computed and shown as a heatmap. The face borders are also depicted to investigate the impact of each part of the face on inpainting the missed eye.

FID and LPIPS are patch-based metrics and do not consider homogeneity and symmetricity in the face. Thus, we propose the symmetry concentration score (SCS) to assess the symmetry of the left and right sides of the face. In Figure 5, we presented heatmaps that depict the influence the model takes from a K×K block of the face during the inpainting of an eye or half of the face. Our symmetry concentration score measures the amount of attention the network pays to a part of the face while repairing the same semantic part on the other side. To acquire this metric, we calculate the mean of the overlapping K × K patches with the desired organs while considering three different patch sizes (16×16, 32×32, 64×64). The results are shown in Table 2. Our method using multiple semantic discriminators performed better than the Swin network without these discriminators. Also, we achieved better symmetric results than [6].

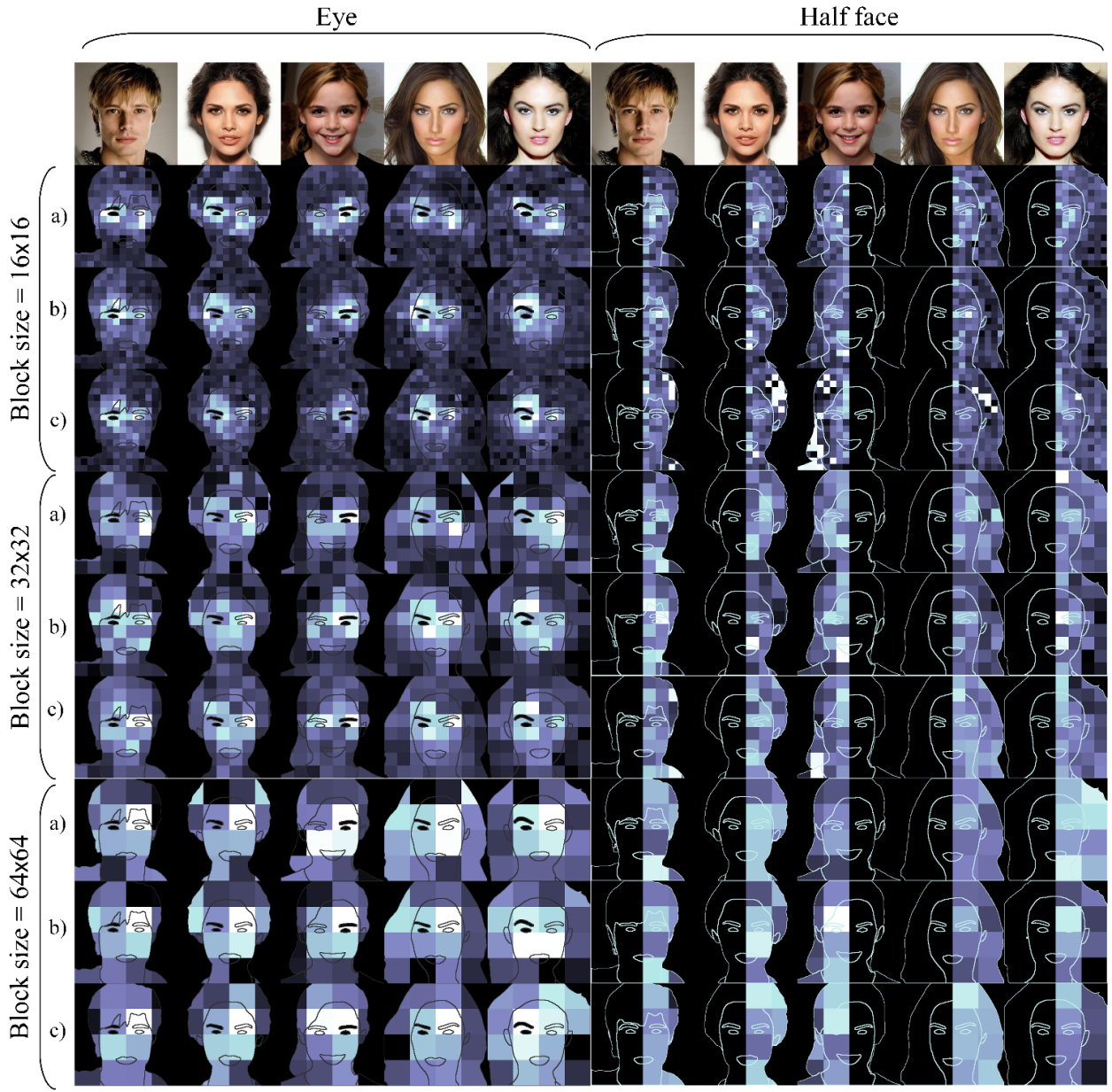

**Fig. 5.** Influence that the generator gets from each K×K block of the face receives to repair the missing parts of the face. a) Lama, b) Swin, c) SFI-Swin. The missing parts are shown as empty black regions (an eye or half of the face), while the background is not considered in these experiments.

Table 1. Presenting the performance of our method SFI-Swin to inpaint different types of masks comparing to powerful methods in this field using two popular metrics, FID and LPIPS. The three best models in each column are shown in red, orange, and green.

| Train masks type | Methods | CelebA-HQ (256×256) | | | | | | | | | | | |
|---|---|---|---|---|---|---|---|---|---|---|---|---|---|
| | | Narrow masks | | | | Medium masks | | | | Wide masks | | | |
| | | 40-50% masked | | All samples | | 40-50% masked | | All samples | | 40-50% masked | | All samples | |
| | | FID | LPIPS | FID | LPIPS | FID | LPIPS | FID | LPIPS | FID | LPIPS | FID | LPIPS |
| Aggressive train masks | Ours (SFI-Swin) | 23.7 | 0.157 | 7.44 | 0.101 | 33.43 | 0.161 | 5.54 | 0.088 | 26.81 | 0.102 | 5.97 | 0.104 |
| | Swin [30] | 23.2 | 0.151 | 7.31 | 0.098 | 32.88 | 0.160 | 5.60 | 0.086 | 26.69 | 0.101 | 5.98 | 0.102 |
| | LaMa-Fourier [6] | 22.7 | 0.132 | 7.26 | 0.085 | 34.1 | 0.145 | 6.13 | 0.080 | 27.8 | 0.168 | 6.96 | 0.098 |
| Narrow train masks | CoModGAN [1] | 35.9 | 0.139 | 16.8 | 0.079 | 48.4 | 0.169 | 19.4 | 0.092 | 64.4 | 0.191 | 24.4 | 0.102 |
| | AOT GAN [2] | 21.0 | 0.127 | 6.67 | 0.081 | 39.1 | 0.162 | 7.28 | 0.089 | 40.4 | 0.204 | 10.3 | 0.118 |
| | RegionWise [3] | 32.5 | 0.188 | 11.1 | 0.124 | 40.4 | 0.179 | 7.52 | 0.101 | 33.9 | 0.205 | 8.54 | 0.121 |
| | DeepFill v2 [5] | 37.0 | 0.201 | 12.5 | 0.190 | 45.3 | 0.189 | 9.05 | 0.105 | 43.0 | 0.214 | 11.2 | 0.126 |
| | EdgeConnect [4] | 29.2 | 0.156 | 9.61 | 0.099 | 40.5 | 0.174 | 7.56 | 0.095 | 34.7 | 0.205 | 9.02 | 0.120 |
| Wide train masks | RegionWise [3] | 47.5 | 0.246 | 17.9 | 0.164 | 50.9 | 0.220 | 10.3 | 0.124 | 42.6 | 0.233 | 11.2 | 0.140 |
| | DeepFill v2 [5] | 30.4 | 0.169 | 9.99 | 0.108 | 40.3 | 0.173 | 7.65 | 0.095 | 34.6 | 0.196 | 8.95 | 0.115 |
| | EdgeConnect [4] | 55.5 | 0.248 | 18.3 | 0.152 | 40.2 | 0.174 | 7.79 | 0.097 | 32.7 | 0.196 | 8.43 | 0.116 |

Table 2. Comparing the symmetry concentration score (SCS) of our method SFI-Swin to inpaint certain parts of the face compared to Swin and LaMa [6].

| Method | Metric + face parts | SCS for eye | SCS for half face |
|---|---|---|
| Ours (SFI-Swin) | **0.7177** | **0.4233** |
| Swin [30] | 0.6319 | 0.3948 |
| LaMa [6] | 0.6225 | 0.3740 |

# 5. Conclusion

This paper discussed the effect of using multiple semantic discriminators incorporated with the Swin transformer-based architecture to repair face images. Our proposed method preserved the symmetry and homogeneity of the face parts. Our experimental results show the proposed method's superiority over powerful rivals, especially on the medium and wide masks.
We also proposed a new method to assess the concentration of the inpainting network while inpainting a specific face organ.
By using multiple discriminators to compute the reality of each facial organ, the generator was guided to preserve the symmetry and homogeneity of the face. This resulted in a generator that resolved one of the most critical inpainting shortcomings.

# References

[1] S. Zhao *et al.*, "Large scale image completion via co-modulated generative adversarial networks," *arXiv Prepr. arXiv2103.10428*, 2021.

[2] Y. Zeng, J. Fu, H. Chao, and B. Guo, "Aggregated contextual transformations for high-resolution image inpainting," *IEEE Trans. Vis. Comput. Graph.*, 2022.

[3] Y. Ma *et al.*, "Region-wise Generative Adversarial Image Inpainting for Large Missing Areas," *IEEE Trans. Cybern.*, 2022.

[4] K. Nazeri, E. Ng, T. Joseph, F. Z. Qureshi, and M. Ebrahimi, "Edgeconnect: Generative image inpainting with adversarial edge learning," *arXiv Prepr. arXiv1901.00212*, 2019.

[5] J. Yu, Z. Lin, J. Yang, X. Shen, X. Lu, and T. S. Huang, "Free-form image inpainting with gated convolution," in *Proceedings of the IEEE/CVF international conference on computer vision*, 2019, pp. 4471–4480.

[6] R. Suvorov *et al.*, "Resolution-robust large mask inpainting with Fourier convolutions," in *Proceedings of the IEEE/CVF Winter Conference on Applications of Computer Vision*, 2022, pp. 2149–2159.

[7] P. Isola, J.-Y. Zhu, T. Zhou, and A. A. Efros, "Image-to-image translation with conditional adversarial networks," in *Proceedings of the IEEE conference on computer vision and pattern recognition*, 2017, pp. 1125–1134.

[8] Z. Liu *et al.*, "Swin transformer: Hierarchical vision transformer using shifted windows," in *Proceedings of the IEEE/CVF International Conference on Computer Vision*, 2021, pp. 10012–10022.

[9] X. Zhang, D. Zhai, T. Li, Y. Zhou, and Y. Lin, "Image inpainting based on deep learning: A review," *Inf. Fusion*, 2022.

[10] Z. Qin, Q. Zeng, Y. Zong, and F. Xu, "Image inpainting based on deep learning: A review," *Displays*, vol. 69, p. 102028, 2021.

[11] J. Jam, C. Kendrick, K. Walker, V. Drouard, J. G.-S. Hsu, and M. H. Yap, "A comprehensive review of past and present image inpainting methods," *Comput. Vis. image Underst.*, vol. 203, p. 103147, 2021.

[12] S. Su, M. Yang, L. He, X. Shao, Y. Zuo, and Z. Qiang, "A Survey of Face Image Inpainting Based on Deep Learning," in *International Conference on Cloud Computing*, 2022, pp. 72–87.

[13] C. Barnes, E. Shechtman, A. Finkelstein, and D. B. Goldman, "PatchMatch: A randomized correspondence algorithm for structural image editing," *ACM Trans. Graph.*, vol. 28, no. 3, p. 24, 2009.

[14] A. Criminisi, P. Pérez, and K. Toyama, "Region filling and object removal by exemplar-based image inpainting," *IEEE Trans. image Process.*, vol. 13, no. 9, pp. 1200–1212, 2004.

[15] J. Yu, Z. Lin, J. Yang, X. Shen, X. Lu, and T. S. Huang, "Generative image inpainting with contextual attention," in *Proceedings of the IEEE conference on computer vision and pattern recognition*, 2018, pp. 5505–5514.

[16] D. Pathak, P. Krahenbuhl, J. Donahue, T. Darrell, and A. A. Efros, "Context encoders: Feature learning by inpainting," in *Proceedings of the IEEE conference on computer vision and pattern recognition*, 2016, pp. 2536–2544.

[17] Z. Zhao *et al.*, "Prior based human completion," in *Proceedings of the IEEE/CVF Conference on Computer Vision and Pattern Recognition*, 2021, pp. 7951–7961.

[18] Z. Hui, J. Li, X. Wang, and X. Gao, "Image fine-grained inpainting," *arXiv Prepr. arXiv2002.02609*, 2020.

[19] Y. Zhou, C. Barnes, E. Shechtman, and S. Amirghodsi, "Transfill: Reference-guided image inpainting by merging multiple color and spatial transformations," in *Proceedings of the IEEE/CVF conference on computer vision and pattern recognition*, 2021, pp. 2266–2276.

[20] A. Dosovitskiy *et al.*, "An image is worth 16x16 words: Transformers for image recognition at

scale," *arXiv Prepr. arXiv2010.11929*, 2020.

[21] L. Yuan *et al.*, "Tokens-to-token vit: Training vision transformers from scratch on imagenet," in *Proceedings of the IEEE/CVF International Conference on Computer Vision*, 2021, pp. 558–567.

[22] H. Yin, A. Vahdat, J. M. Alvarez, A. Mallya, J. Kautz, and P. Molchanov, "A-ViT: Adaptive Tokens for Efficient Vision Transformer," in *Proceedings of the IEEE/CVF Conference on Computer Vision and Pattern Recognition*, 2022, pp. 10809–10818.

[23] K. Han *et al.*, "A survey on vision transformer," *IEEE Trans. Pattern Anal. Mach. Intell.*, 2022.

[24] Z. Wan, J. Zhang, D. Chen, and J. Liao, "High-fidelity pluralistic image completion with transformers," in *Proceedings of the IEEE/CVF International Conference on Computer Vision*, 2021, pp. 4692–4701.

[25] C. Zheng, T.-J. Cham, and J. Cai, "Tfill: Image completion via a transformer-based architecture," *arXiv Prepr. arXiv2104.00845*, 2021.

[26] W. Li, Z. Lin, K. Zhou, L. Qi, Y. Wang, and J. Jia, "MAT: Mask-Aware Transformer for Large Hole Image Inpainting," in *Proceedings of the IEEE/CVF Conference on Computer Vision and Pattern Recognition*, 2022, pp. 10758–10768.

[27] Q. Dong, C. Cao, and Y. Fu, "Incremental transformer structure enhanced image inpainting with masking positional encoding," in *Proceedings of the IEEE/CVF Conference on Computer Vision and Pattern Recognition*, 2022, pp. 11358–11368.

[28] J. Wang, S. Chen, Z. Wu, and Y.-G. Jiang, "FT-TDR: Frequency-guided Transformer and Top-Down Refinement Network for Blind Face Inpainting," *IEEE Trans. Multimed.*, 2022.

[29] L. Chi, B. Jiang, and Y. Mu, "Fast Fourier convolution," *Adv. Neural Inf. Process. Syst.*, vol. 33, pp. 4479–4488, 2020.

[30] H. Cao *et al.*, "Swin-unet: Unet-like pure transformer for medical image segmentation," *arXiv Prepr. arXiv2105.05537*, 2021.

[31] P. Wang, Y. Li, and N. Vasconcelos, "Rethinking and improving the robustness of image style transfer," in *Proceedings of the IEEE/CVF Conference on Computer Vision and Pattern Recognition*, 2021, pp. 124–133.

[32] C. Yu, C. Gao, J. Wang, G. Yu, C. Shen, and N. Sang, "Bisenet v2: Bilateral network with guided aggregation for real-time semantic segmentation," *Int. J. Comput. Vis.*, vol. 129, no. 11, pp. 3051–3068, 2021.

[33] T. Karras, T. Aila, S. Laine, and J. Lehtinen, "Progressive growing of gans for improved quality, stability, and variation," *arXiv Prepr. arXiv1710.10196*, 2017.

[34] R. Zhang, P. Isola, A. A. Efros, E. Shechtman, and O. Wang, "The unreasonable effectiveness of deep features as a perceptual metric," in *Proceedings of the IEEE conference on computer vision and pattern recognition*, 2018, pp. 586–595.

[35] M. Heusel, H. Ramsauer, T. Unterthiner, B. Nessler, and S. Hochreiter, "Gans trained by a two time-scale update rule converge to a local nash equilibrium," *Adv. Neural Inf. Process. Syst.*, vol. 30, 2017.

[36] A. Paszke *et al.*, "Pytorch: An imperative style, high-performance deep learning library," *Adv. Neural Inf. Process. Syst.*, vol. 32, 2019.

[37] D. P. Kingma and J. Ba, "Adam: A method for stochastic optimization," *arXiv Prepr. arXiv1412.6980*, 2014.